\documentclass{article}

\usepackage{arxiv}

\usepackage[utf8]{inputenc} 
\usepackage[T1]{fontenc}    
\usepackage{hyperref}       
\usepackage{url}            
\usepackage{booktabs}       
\usepackage{amsfonts}       
\usepackage{nicefrac}       
\usepackage{microtype}      

\usepackage{lipsum}
\usepackage{graphicx}
\usepackage{multirow}
\usepackage{color,array,dcolumn}
\usepackage{amsmath}
\usepackage{amssymb}
\usepackage[export]{adjustbox}
\usepackage{bbm}

\newcolumntype{L}[1]{>{\raggedright\let\newline\\\arraybackslash\hspace{0pt}}m{#1}}
\newcolumntype{C}[1]{>{\centering\let\newline\\\arraybackslash\hspace{0pt}}m{#1}}
\newcolumntype{R}[1]{>{\raggedleft\let\newline\\\arraybackslash\hspace{0pt}}m{#1}}

\title{Multimodal learning-based inversion models for the space-time reconstruction of satellite-derived geophysical fields}
\author{
  R. Fablet\thanks{Corresponding author} \\
  IMT Atlantique, Lab-STICC,
  Brest, FR \\
  \texttt{ronan.fablet@imt-atlantique.fr} \\
   \And
  B. Chapron \\
  Ifremer, LOPS,
  Brest, FR \\
  \texttt{bertrand.chapron@ifremer.fr} \\}

\begin{document}
\maketitle

\begin{abstract}
Satellite altimetry provides a direct observation of the divergence-free component of sea surface currents. The associated space-time sampling however prevents from retrieving fine-scale sea surface dynamics, typically below a 10-day time scale. By contrast, other satellite sensors provide higher-resolution observations of sea surface tracers such as sea surface temperature (SST). 
Under specific dynamical regimes such as Surface Quasi-Geostrophic dynamics, SSH-SST synergies enhance the 
reconstruction of sea surface dynamics. The generalization to other dynamical regimes remain a challenge. Here, we investigate this issue from a physics-informed learning perspective. We introduce a trainable multimodal inversion schemes for the reconstruction of sea surface dynamics from satellite-derived SSH and SST observations. The proposed approach combines a variational formulation with trainable observation and {\em a priori} terms with a trainable solver. An observing system simulation experiment for a Gulf Stream with nadir along-track and wide-swath altimetry data supports the relevance of our approach compared with state-of-the-art schemes. We report relative improvement of 40\% to 50\% w.r.t. the operational product in terms of root mean square error and resolved space-time scales. 
\end{abstract}
\section{Introduction}
\label{sec:intro}

\begin{figure}[tbh]
    \centering
    \includegraphics[width=8.25cm]{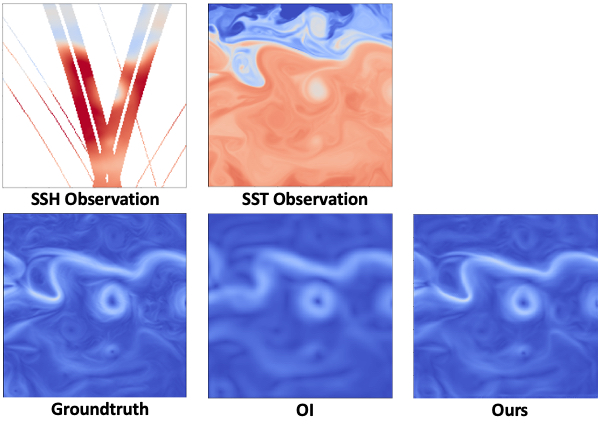}
    \caption{{\bf Multimodal reconstruction of sea surface height (SSH)}: first row, SSH  and SST (sea surface temperature) observation; second row, SHH gradient magnitude fields for the true field, an optimal interpolation using only SSH data \cite{taburet_duacs_2019} and the proposed multimodal framework.}
    \label{fig:res1}
\end{figure}

Satellite earth observation involves a variety of sensors which routinely deliver observation data. These sensors differ in their observation process (e.g., infrared sensors, SAR imaging, wide-swath vs. nadir along-track sensor,...) as well as in their sensitivity to the atmospheric conditions. The retrieval of satellite-derived geophysical fields (e.g., sea surface parameters, land surface parameters, rainfall, wind,...) could then benefit from multimodal synergies. 
 As an illustration, we depict in Fig.\ref{fig:res1} an example for sea surface parameters. While sea surface height (SSH) is scarcely observed by satllite altimeters, sea surface temperature snapshots from infrared and microwave sensors offer a much better space-time sampling. As sea surface height (SSH) and temperature (SST) fields clearly share common features due to the underlying ocean dynamics, one would expect to benefit from the availability of SST data to enhance the reconstruction of the SSH. The derivation of a physical law which could relate these two fields is however far from being straightforward. Despite theoretical advances and numerical studies \cite{isern-fontanet_potential_2006,rio_improving_2016}, existing models are restricted to specific dynamical regimes and do not generalize well.

In this context, end-to-end learning-based inversion schemes appear as highly appealing solutions to address multimodal reconstruction issues \cite{dieleman_end--end_2014,schwartz_deepisp_2019} as they might directly learn some underlying multimodal representations \cite{ngiam_multimodal_2011}. Due to the irregularly sampling of the observations with very high missing data rates \cite{gaultier_challenge_2015}, image-to-image translation schemes \cite{zhu_toward_2017} do ot however apply. And recent advances in physics-informed learning \cite{fablet_learning_2021} for inverse problems seem more relevant to jointly benefit from some physical knowledge and the efficiency of learning approaches.

Here, we extend \cite{fablet_learning_2021} to a novel multimodal learning-based inversion model. The proposed end-to-end learning scheme combines a variational formulation with trainable observation and {\em a priori} terms with a trainable solver. We show that we can jointly learn from data both the multimodal observation model, the prior and the solver. We demonstrate the relevance of the proposed approach through an application to the reconstruction of sea surface dynamics from irregularly-sampled SST-SSH satellite image time series.


\section{Problem statement and Related work}
\label{sec:format}

The reconstruction of image time series of geophysical parameters from satellite observation data is referred to as a data assimilation problem in the geoscience literature \cite{evensen_data_2009}. It generally follows a model-driven approach according to a state-space formulation
\begin{equation}
\label{eq: state space}
\left \{\begin{array}{ccl}
    \displaystyle \frac{\partial x(t)}{\partial t} &=& {\cal{M}}\left (x(t) \right )+ \eta(t)\\~\\
    y_m(t) &=& {\cal{H}}_m\left ( x(t) \right ) + \epsilon_m(t), \forall t \\
\end{array}\right.
\end{equation}
where $x$ is the space-time process of interest defined over a space domain ${\cal{D}}$ and a time interval $[0,T]$. $x$ is governed by dynamical model ${\cal{M}}$. $y_m$ is the space-time observation data for observation modality $m$. Observation operator ${\cal{H}}_m$ relates $x$ to $y_m$. $\eta$ and $\{\epsilon_m\}_m$ refer to noise processes.

The reconstruction of state sequence $x$ from given observation data $\{y_k(t_i)\}_{i,k}$ at time steps $\{t_i\}$ may then be stated as the minimization of a variational criterion
\begin{equation}
\label{eq:4dvar model}
\begin{array}{ccl}
\displaystyle U_\Phi\left ( x , \{y_m\} \right ) &=& \displaystyle \sum_m \lambda_{m} \sum_i \left \| y_m(t_i)-{\cal{H}}_m \left ( x(t_i) \right )\right \|^2 \\~\\
&+& \displaystyle\gamma \sum_i \left \|x(t_i) - \Phi(x)(t_i) \right \|^2
\end{array}
\hspace*{-0.5cm}
\end{equation}
where $\Phi(x)(t_i)$ is the time integration of dynamical model  ${\cal{M}}$ over time interval $[t_{i-1},t_i]$.  $\{\lambda_{m}\}_m$ and $\gamma$ are the weighing parameters of the observation and prior terms. 

Depending on the parameterizations for observation models 
$\{{\cal{H}}_m\}_m$ and dynamical model ${\cal{M}}$, different optimization algorithms may be considered for the above minimization. Especially, when considering numerical implementations in deep learning and differentiable frameworks (e.g., pytorch), one may benefit from the embedded automatic differentiation tools to implement gradient descent algorithms. This also opens avenues for considering pre-trained operators, which have recently been explored for computational imaging problems both for plug-and-play priors and pre-trained observation operators \cite{zhang_plug-and-play_2021,wei_tuning-free_2020}. We may point out that in such schemes, there is no guarantee for the pre-trained operators to be fully-relevant for the considered inversion tasks. 

End-to-end learning approaches may address these limitations as one aims at learning an inverse model using some reconstruction performance metrics as training loss \cite{aggarwal_modl_2019,lucas_using_2018,gastineau_generative_2022}. 
In this line of research, one may design physics-informed end-to-end learning approaches, which explicitly rely on variational formulations \cite{fablet_learning_2021,kobler_total_2020}. Here, we further extend our prior work \cite{fablet_learning_2021} to multimodal inversion problems.

\section{Proposed approach}
\label{sec:pagestyle}

This section introduces the proposed multimodal learning-based inversion scheme. 

\subsection{Multimodal trainable variational formulation}
\label{ss:4dvar model}
Without loss of generality, we derive from (\ref{eq:4dvar model}) the following matrix-based variational formulation 
\begin{equation}
\label{eq:4dvar model 2}
\begin{array}{ccl}
\displaystyle U\left ( x , \{y_m\} \right ) &=& \displaystyle \sum_{m,n} \lambda_{m,n} \left \| {\cal{G}}_{m,n} \left ( y_m , x\right ) \right \|^2 \\~\\
&+& \displaystyle\gamma \left \|x - \Phi(x) \right \|^2
\end{array}
\end{equation}
where both $\{{\cal{G}}_{m,n}\}_{m,n}$ and $\Phi$ are trainable operators implemented as neural networks.

Compared with (\ref{eq:4dvar model}), we rewrite the observation terms according to operators ${\cal{G}}_{m,l}$. They aim at extracting relevant signatures from observation $y_m$ which inform state $x$. We may recover formulation (\ref{eq:4dvar model}) when ${\cal{G}}_{m,n} \left ( y_m, x\right ) = y_m -{\cal{H}}_{m}(x)$. It embeds advection operator if observation $y_k$ is regarded as a passive tracer transported by some dynamics governed by state $x$
\begin{equation}
\label{eq: advection}
    {\cal{G}}_{m,n} \left ( y_m , x\right ) = \frac{\partial d y_m}{\partial t} - \langle \nabla y_m , V\left (x\right) \rangle 
\end{equation}
with $V (x)$ the velocity field associated with state $x$.
We can also explore the existence of similar features shared by  observation $y_k$ and state $x$ using the following parameterization
\begin{equation}
\label{eq: multimodal obs}
    {\cal{G}}_{m,n} \left ( y_m , x\right ) = {\cal{G}}^1_{m,n} \left ( y_m \right) - {\cal{G}}^2_{m,n} \left ( x \right)
\end{equation}
with ${\cal{G}}^1_{k,l}$ and ${\cal{G}}^2_{k,l}$ trainable operators acting as feature extraction operators in (\ref{eq:4dvar model 2}).

Operator $\Phi$ states the prior onto the state sequence to be reconstructed. While in a model-driven setting it derives from known governing equations for state $x$, our previous work suggests that the parameterization of $\Phi$ using state-of-the-art neural network architectures such as U-Nets might lead to better inversion performance as the form of the prior can adapt to the considered inversion problem and observation patterns. Here, we follow the latter approach and consider a two-scale U-Net with bilinear units as in \cite{fablet_learning_2021}.

\subsection{End-to-end inversion model}

Based on the variational formulation introduced in the previous section, we derive an end-to-end inversion model given by a trainable gradient-based iterative scheme \cite{fablet_learning_2021} according to the following update rule at iteration $k$:
\begin{equation}
\label{eq: lstm update}
\left \{\begin{array}{l}
     g^{(k+1)} =   LSTM \left[ \cdot \nabla_x U \left ( x^{(k)},\{y_m\} \right),  h(k) , c(k) \right ]  \\~\\
     x^{(k+1)} = x^{(k)} - {\cal{L}}  \left( g^{(k+1)} \right )  \\
\end{array} \right.\hspace*{-0.2cm}
\end{equation} where $\{h(k) , c(k)\}$ are the internal states of the LSTM model. ${\cal{L}}$ is a linear layer to map the LSTM output to the space spanned by state $x$. 

Overall, the resulting end-to-end architecture uses as inputs observation data $\{y_m\}$ and some state initialisation $x^{(0)}$ to output the reconstructed state. It implements a predefined number of the above gradient steps. Let us denote by $\widehat{x}=\Psi _{\Theta} \left ( x^{(0)} , \{y_m\} \right )$ the output of the end-to-end inversion scheme where $\Theta$ stands for the set of all trainable parameters, which comprise those of prior $\Phi$,  observation operators ${\cal{G}}_{m,n}$ and LSTM-based iterative update.

\subsection{Learning scheme}

We exploit a supervised learning strategy. Given a training dataset comprising triplets of true states, observation data and initial conditions $\{x^{true}_i,\{y_{i,m}\},x^{(0)}_i\}_i$, the training loss typically involves a weighted sum of the reconstruction error for state $x$ and its gradient
\begin{equation}
{\cal{L}}_{x} =  \sum_i \left \| x^{true}_i - \widehat{x}_i \right \|^2
\end{equation}
\begin{equation}
{\cal{L}}_{\nabla x} = \sum_i \left \| \nabla x^{true}_i - \nabla\widehat{x}_i \right \|^2
\end{equation}
Following \cite{fablet_learning_2021}, we consider the following regularisation terms
\begin{equation}
{\cal{L}}_{\Phi} =  \sum_i \left \| x^{true}_i - \Phi \left ( x^{true}_i  \right ) \right \|^2+  \sum_i \left \| \widehat{x}_i - \Phi \left ( \widehat{x}_i \right ) \right \|^2
\end{equation}
The overall training loss is a weighted sum of these different terms. The training scheme uses Adam optimizer with a decaying learning rate and an increasing number of gradient-based iterations in the trainable solver. The Pytorch code of our implementation is available at \url{https://github.com/CIA-Oceanix/4dvarnet-core}.  

\section{Application to satellite-derived sea surface dynamics}

This section details the considered application to the reconstruction of image time series of sea surface dynamics from irregularly-sampled multi-source satellite data. 

\subsection{Dataset and experimental setting}


We consider a multi-source setting using sea surface height (SSH) and sea surface temperature (SST). As illustrated in Fig.\ref{fig:res1}, the gradient of the SSH provides an estimate of sea surface currents, as the sea surface height relates to the divergence-free component of the sea surface current \cite{taburet_duacs_2019,ubelmann_dynamic_2014}. The considered SSH data combine nadir altimeters and upcoming wide-swath SWOT altimeters \cite{taburet_duacs_2019,gaultier_challenge_2015}. The associated sampling is strongly irregular and cover about 5\% of the considered domain on a daily time scale. We assume gap-free observations of daily sea surface temperature (SST) fields, which may be retrieved from microwave satellite sensors \cite{isern-fontanet_potential_2006}. 

As case-study, we rely on the benchmarking framework presented in \cite{le_guillou_mapping_2020}. It relies on a high-resolution numerical simulation of ocean dynamics and the simulation of real satellite observation patterns. This dataset involves a one-year time series of 200x200 daily images within a 10$^\circ$x10$^\circ$ area along the Gulf Stream from October 2018 to September 2019. We refer the reader to the following link \footnote{\url{https://github.com/ocean-data-challenges/2020a_SSH_mapping_NATL60}} for a detailed presentation of the dataset. The performance metrics \cite{le_guillou_mapping_2020} are as follows: $\mu$, the normalized root-mean-square-error-based metrics equals 1 for a perfect reconstruction; $\sigma$, the standard deviation of the above RMSE score; $\lambda_t$, the minimum time scale resolved in days; $\lambda_x$, the minimum spatial scale resolved in degrees. The last two metrics are computed in the spectral domain. We evaluate all performance metrics for a 40-day period from October, 22 2018 to December, 2 2018. We use as training dataset the data from February 2019 to September 2019 and as validation dataset data from January 2019.  

\begin{table*}[tb]
    \footnotesize
    \centering
    \begin{tabular}{|C{2.cm}|C{1.9cm}|C{2cm}|C{1.75cm}|C{1.75cm}|C{1.75cm}|}
    \toprule
    \toprule
     \bf Approach& \bf Data sources&\bf $mu$ &$\sigma$ & $\lambda_x$ ($^\circ$)& $\lambda_t$ (days)\\
    \toprule
    \toprule
     DUACS \cite{taburet_duacs_2019} & SSH only& 0.92 & 0.02 & 1.22 & 11.15 \\    
     DYMOST \cite{ubelmann_dynamic_2014} & SSH only& 0.93 & 0.02 & 1.20 & 10.07  \\    
     MIOST \cite{ubelmann_reconstructing_2021} & SSH only& 0.94 & 0.01 & 1.18 & 10.14 \\    
     BFN \cite{le_guillou_mapping_2020} & SSH only & 0.93 & 0.02 & 0.8 & 10.09  \\    
    \toprule
     U-Net & SSH only & 0.93 & 0.01 & 1.19 & 9.96 \\
      & SSH and SST & 0.95 & 0.01 & 7.56 & 5.96 \\
    \toprule
     4DVarNet (ours) & SSH only & 0.95 & 0.01 & 0.82 & 6.57  \\
     & SSH and SST & \bf 0.96 & \bf 0.01& \bf 0.64& \bf 4.08\\
    \bottomrule
    \bottomrule
    \end{tabular}
    \caption{{\bf Synthesis of reconstruction performance:} we report the performance metrics of the benchmarked approaches for the reconstruction of image time series of sea surface currents from satellite data. We refer the reader to the main text for the description of the different metrics. We highlight in bold the best score. }
    \label{tab:res}
\end{table*}

\subsection{Proposed framework}

We detail in this section the considered parameterization of the proposed framework, referred to as 4DVarNet. 

{\bf Observation and state variables:} We consider three observation data sources: nadir altimeter and SWOT SSH data $y_1$, optimally-interpolated SSH fields $y_2$ using  $y_1$ and SST data $y_3$. We denote by $\Omega$ the mask for altimetry observations and $\mathbbm{1}_\Omega$ the associated indicator function. State $x$ involves the following decomposition, $x=(\overline{x},\delta x)$, such that $\overline{x}+\delta x$ is the SSH field. $\overline{x}$ refers to a large-scale component and $\delta x$ to a fine-scale one.

{\bf Observation terms:} For SSH data sources, $m=1,2$, we consider identity observation operators in (\ref{eq:4dvar model 2}) such that: ${\cal{G}}_{1,1} \left ( y_1 , x\right ) = \mathbbm{1}_\Omega\cdot(y_1-\overline{x}-\delta x)$ 
and ${\cal{G}}_{2,1} \left ( y_2 , x\right ) = y_2-\overline{x}$. Regarding SST data source, $m=3$, we explore whether local features may be shared by the SSH and SST fields as evidenced in the SQG (Surface Quasi-Geostrophic) dynamical mode \cite{isern-fontanet_potential_2006}. This leads to parameterization (\ref{eq: multimodal obs}) for observation operators $\{{\cal{G}}_{3,k}\}_k$. Here, we investigate convolution operators for ${\cal{G}}^1_{3,k}$ and ${\cal{G}}^2_{3,k}$. Numerical experiments for more complex non-linear operators did not bring significant improvement, neither using advection-based operator (\ref{eq: advection})  

{\bf Dynamical prior:} Regarding prior $\Phi$ in (\ref{eq:4dvar model 2}), we follow a parameterization similar to that considered in \cite{fablet_learning_2021} and use a two-scale U-Net \cite{cicek_3d_2016} with bilinear blocks to account for the non-linearities expected in upper ocean dynamics, especially advection-related processes. We consider 7-day time windows such that state $x$ is a 7x200x200x2 tensor. 

{\bf Trainable solver:} Regarding the trainable LTSM-based solver, we consider a 2{\sc d} convolutional LSTM with 100-dimensional hidden states.  

Overall, our end-to-end architecture involves 1,400,000 parameters, with  925,000 for the trainable solver. 

\subsection{Results}

We report in Tab.\ref{tab:res} the synthesis of the considered benchmarking experiments. We compare the proposed approach with and without SST data to state-of-the-art data assimilation schemes \cite{le_guillou_mapping_2020}: namely the operational optimal-interpolation-based method (DUACS) \cite{taburet_duacs_2019}, model-driven interpolations \cite{le_guillou_mapping_2020,ubelmann_dynamic_2014} and multiscale interpolation \cite{ubelmann_reconstructing_2021}. We also report the performance of a direct U-Net-based direct inversion scheme using a U-Net architecture similar to that of operator $\Phi$ in our approach (cf. (\ref{eq:4dvar model 2})). The proposed multimodal scheme significantly outperforms all the other ones, especially regarding the resolved spatial and temporal scales. For instance, we report a relative gain in terms of RMSE of \~40\% w.r.t. 
DUACS scheme, which is the operational baseline.
We may point out that compared with schemes from the space oceanography literature, only our method retrieves temporal scales below 7 days, whereas the other approaches only resolve time scales up to \~10 days. Interestingly, we also significantly outperform a direct U-Net-based inversion scheme. While the RMSE score of this direct inversion improves with SSH-SST data sources, this multimodal setting may also generate local artifacts stressed by the poor resolved spatial scale above 2$^\circ$.
These results emphasize the role of the trainable LSTM solver, which seems particularly noticeable to ensure the reconstruction of spatially-consistent SSH fields with a resolved spatial scale up to 0.64$^\circ$. These results also evidence the relevance of SST-SSH synergies to enhance the reconstruction of SSH fields: for instance with a significant improvement of the resolved spatial and time scales, 0.82$^\circ$ vs 0.64$^\circ$ for $\lambda_x$, and 6.57days vs. 4.08days for $\lambda_t$.

\begin{figure}[htb]
    \centering
    \includegraphics[width=7cm]{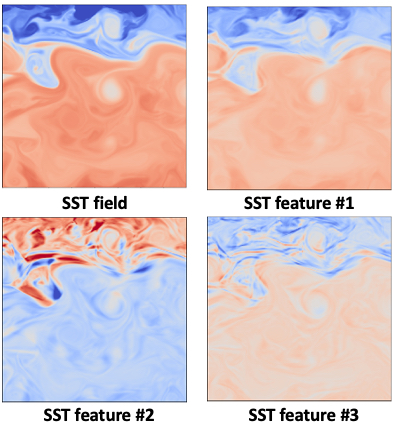}
    \caption{{\bf Learnt SST features:} we depict the maps of the learnt features extracted from SST images according to operators $\{{\cal{G}}_{3,l}\}_l$ to inform sea surface dynamics from SST fields.}
    \vspace*{-0.5cm}
    \label{fig:sst features}
\end{figure}

Visually, we emphasize in Fig.\ref{fig:res1} the clear improvement w.r.t. the operational processing and the ability to recover most fine-scale structures of the SSH fields. We also report in  Fig.\ref{fig:sst features} different feature maps extracted from the SST field (upper left) with a view to reconstructing the SSH field. They reveal different space-time scales. This may be regarded as a generalization of SGQ dynamical mode for sea surface dynamics, where SST and SSH fields relate up to fractional Laplacian operator \cite{isern-fontanet_potential_2006}. 


\section{Conclusion}

This paper introduced a novel multimodal physics-informed learning-based inversion framework. Through an application to the spatio-temporal reconstruction of the sea surface dynamics from irregularly-sampled satellite-derived SSH and SST data, we demonstrated its ability to learn relevant features which enhance the reconstruction of fine-scale patterns of partially-observed image data. 

Future work may further explore the parameterization of the considered multimodal operators as well as other applications to multimodal satellite data sources (ocean colour, hyperspectral imaging).
  
\section*{\bf Acknowledgements}
This work was supported by LEFE program (LEFE MANU project IA-OAC), CNES (grant OSTST DUACS-HR) and ANR Projects Melody and OceaniX. It benefited from HPC and GPU resources from Azure (Microsoft Azure grant) and from GENCI-IDRIS (Grant 2020-101030).  

\bibliographystyle{plain}

\end{document}